\documentclass[10pt,twocolumn,letterpaper]{article}

\usepackage{iccv}
\usepackage{times}
\usepackage{epsfig}
\usepackage{graphicx}
\usepackage{amsmath}
\usepackage{amssymb}
\usepackage{bbding}

\usepackage{url}
\usepackage{multirow}
\usepackage{makecell}
\usepackage[font=small]{caption}
\usepackage[tight,footnotesize]{subfigure}
\usepackage[utf8]{inputenc}
\usepackage{amsthm}
\usepackage{booktabs}
\usepackage{enumitem}
\usepackage{pifont}
\usepackage{color}
\usepackage[table]{xcolor}

\newcommand{\cmark}{\ding{51}}%
\newcommand{\xmark}{\ding{55}}%

\RequirePackage{algorithm}
\RequirePackage{algorithmic}

\definecolor{mygreen}{RGB}{0,100,0}
\definecolor{myblue}{RGB}{0,0,230}
\definecolor{myred}{RGB}{150,0,0}

\newcommand{\baseliness}{{{manually-designed}}}
\newcommand{\myPara}[1]{\vspace{.05in}\noindent\textbf{#1}}

\usepackage{hyperref}
\hypersetup{pagebackref=false,colorlinks=true,linkcolor=myred,citecolor=myblue,bookmarks=false,urlcolor=black}

\iccvfinalcopy 

\def\iccvPaperID{4044} 

\ificcvfinal\fi

\begin{document}


\title{AutoSpace: Neural Architecture Search with Less Human Interference}


\author{Daquan Zhou\textsuperscript{1},
Xiaojie Jin \textsuperscript{2},
Xiaochen Lian\textsuperscript{2},
Linjie Yang\textsuperscript{2},
Yujing Xue\textsuperscript{1},
Qibin Hou\textsuperscript{1}{$\thanks{Corresponding author.}$},
Jiashi Feng\textsuperscript{1} \\
\textsuperscript{1}National University of Singapore,\textsuperscript{2}ByteDance US AI Lab  \\
\texttt{\small {\{zhoudaquan21, xjjin0731, lianxiaochen, yljatthu, andrewhoux\}}@gmail.com}
\\
\texttt{\small xueyj14@outlook.com, elefjia@nus.edu.sg}
}

\maketitle
\ificcvfinal\thispagestyle{empty}\fi

\begin{abstract}
Current neural architecture search (NAS) algorithms still require expert knowledge and effort to design a search space for network construction.
In this paper, we consider automating the search space design to minimize human interference, which however faces two challenges:
the explosive complexity of the exploration space and the expensive computation cost to evaluate the quality of different search spaces.
To solve them, we propose a novel differentiable evolutionary framework named \emph{AutoSpace}, 
which evolves the search space to an optimal one with following novel techniques:
a differentiable fitness scoring function  to efficiently evaluate the performance of cells and 
a reference architecture  to speedup the evolution procedure and avoid falling into sub-optimal solutions.
The framework is generic and compatible with additional computational constraints, making it feasible to learn specialized search spaces that fit different computational budgets.
With the learned search space, the performance of recent NAS algorithms can be improved significantly compared with using previously manually designed spaces.
Remarkably, the models generated from the new search space achieve 77.8\% top-1  accuracy  on ImageNet under the mobile  setting (MAdds$\leq$500M), outperforming previous SOTA EfficientNet-B0 by 0.7\%. All codes will be made public.
\end{abstract}

\section{Introduction} \label{sec:introduction}

Recently neural architecture search (NAS) algorithms are popularly explored and applied, yielding several state-of-the-art (SOTA) deep neural network architectures~\cite{tan2019efficientnet,tan2019mnasnet,howard2019searching,wan2020fbnetv2}. 
Applying a NAS algorithm typically comprises three steps:
(1) designing a search space by specifying its elementary operators; 
(2) developing a searching algorithm to explore the space and select operators from it to build the candidate model;
and (3) implementing an evaluation strategy to validate the performance of searched models. 
Extensive studies have been devoted to the latter two steps, i.e. searching algorithms and evaluation strategies~\cite{tan2019mixconv,cai2018proxylessnas,howard2019searching,tan2019mnasnet,wu2019fbnet,wan2020fbnetv2,tan2019mixconv,guo2019single,tan2019efficientnet}. 

\begin{figure}[t]
    \centering
    \includegraphics[width=0.48\textwidth]{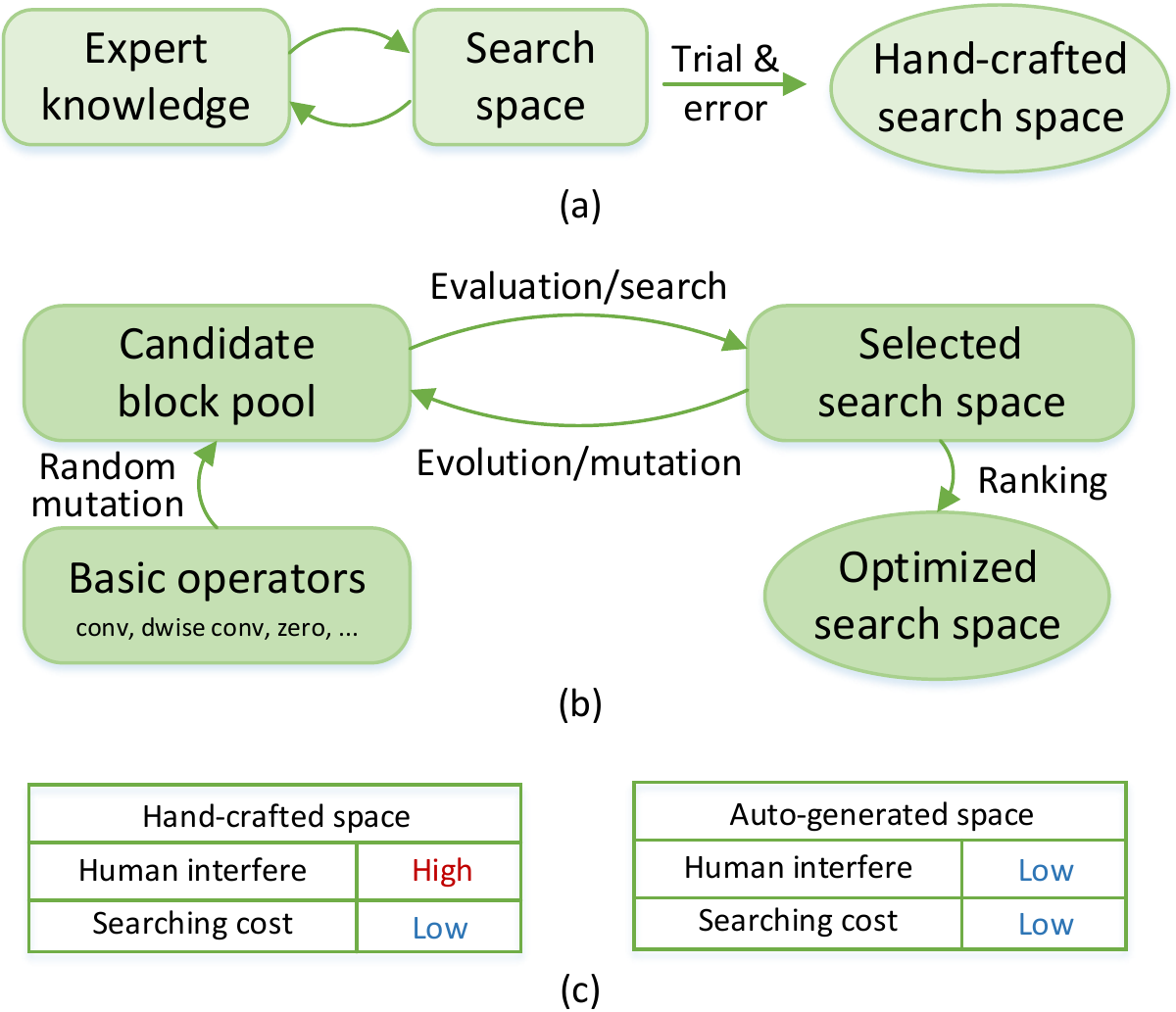}
    \caption{\small \textbf{Comparison of different search space construction schemes.} 
    (a) Most existing NAS methods deploy handcrafted search spaces whose construction heavily relies on expertise and trial-and-error.
    (b) Our proposed method automatically builds and optimizes the search space by learning to form the basic operators into candidate building blocks and using an efficient approach to evolve and evaluate these building blocks. 
    (c) Compared with existing schemes, our proposed one involves lower human effort and searching cost. 
    }
    \label{fig:strategy_compare}
\end{figure}


To reduce the complexity of the search space to explore, 
the common practice of recent NAS algorithms is to leverage human prior knowledge 
to design smaller search spaces, most of which are based on well performing handcrafted building blocks and their variants, \eg, the inverted residual block~\cite{cai2018proxylessnas,tan2019efficientnet,tan2019mixconv,tan2019mnasnet,hou2021coordinate} and the channel shuffling block~\cite{guo2019single}. 
On one hand, using restricted search spaces indeed enables NAS algorithms to enjoy higher efficiency;
however on the other hand, due to such heavy human interference, the possibility to discover novel and better architectures is limited~\cite{elsken2018neural}.
How to reduce human efforts in designing the search space and make the procedure automatic is still under-explored \cite{ren2020comprehensive,surve2018approach}.

In this work, we consider the open architecture space that consists of only basic operators with minimal human prior knowledge on cell graph typologies. 
Despite some early pilot investigations~\cite{zoph2016neural,real2017large}, the progress is much hindered by several practical challenges.
First of all, searching over the open space is unaffordably time consuming due to the combinatorial nature of the problem \cite{ren2020comprehensive}. 
How to fast prune the unnecessary combinations of operators and lower the number of possible candidates to explore is thus necessary but remains an open question.
Secondly, applying the RL-alike algorithms to search from scratch usually suffers poor exploration performance as they easily get stuck at sub-optimum. 
As a result, the model searched from the obtained space may perform no better than the ones from a manually designed search space.


In view of the above challenges, we then wonder whether it is possible to maximally reduce human interference in search space construction in a way such that the algorithms can effectively explore over  the large space within an acceptable time and computation cost budget?
To this end, we develop a novel differentiable AutoSpace framework to automatically evolve the full search space to an optimal subspace for the target applications.
Our main insight is that a one-time searching for an optimal subspace at first and further performing NAS within it
would provide higher exploration capability and searching efficiency at the same time, 
while avoiding getting stuck at the sub-optimum.
Figure~\ref{fig:strategy_compare} illustrates the differences between our space evolving strategy and the previous ones for designing search spaces.

Concretely, AutoSpace starts with an open space that comprises all the possible combinations of basic operators (\eg convolution, pooling, identity mapping). 
Then a differentiable evolutionary algorithm (DEA) is developed to evolve the search space to a subspace of high-quality cell structures.
The subspace can then be adopted in any NAS algorithms seamlessly to find the optimal model architectures. 
To reduce the potential high cost of subspace searching and evaluation, AutoSpace introduces a couple of new techniques to remove the redundant cell structures and improve the parallelism of the evolution process as detailed in Section \ref{sec:method}.  

We verify the superiority of the search space from AutoSpace on the ImageNet~\cite{krizhevsky2012imagenet} dataset.
With the same NAS algorithm, AutoSpace provides  much more accurate models  than the previous SOTA models   searched from the manually designed   spaces. Besides, by combining the cell structures discovered with AutoSpace to EfficientNet, we successfully improve the Top-1 accuracy on ImageNet by 0.7\%.
%

In summary, we make the following contributions:
\begin{itemize}
\item We are among the first to explore the automatic learning of search spaces in NAS algorithms. 
Compared to searching for network architectures on a manually designed search space, searching for a search space is more challenging due to the larger exploration space/computational complexity.
    
    \item We propose a novel learning framework that takes advantage of both the high exploration capability of evolutionary algorithms and the high optimization efficiency of gradient descend methods. 
    The proposed framework can be seamlessly integrated with popular neural architecture searching algorithms.
    
   
    
    \item By directly replacing the original search space with the learned search space, the top-1 classification accuracy of previous SOTA NAS algorithms can be improved significantly at different model sizes. Specifically, at 200M MAdds, the performance of the searched model is improved by more than 1.8\% on ImageNet. 
    
\end{itemize}

\section{Related Work} 

\label{sec:related_work}

Most of the previous neural architecture search (NAS)  works focus on better searching algorithms while the design of the search space is less studied.
This is primarily due to the unaffordable computation cost for automatically searching a search space. 
For example, a recent method of evaluating search spaces uses empirical distribution function (EDF)~\cite{radosavovic2019network} where each evaluation iteration  takes 25k GPU hours\footnote{The work \cite{radosavovic2019network} evaluates the distribution on 50k models and the reported training speed is 2 models per GPU hour.} even on a small dataset of CIFAR10~\cite{krizhevsky2009learning}.

Thus, most of the NAS algorithms use manually designed search spaces. 
An early work NAS-RL~\cite{zoph2016neural} defines its space via macro and micro architectures.
The macro architecture is used for connections between layers and the micro architecture space includes the structural hyper-parameters for each filter within a layer.
Such a huge space is extremely hard or even impractical to enumerate each candidate within it for evaluation. 
Thereafter, most NAS methods change to adopt size-reduced  search spaces to improve their searching efficiency.
For example, the cell based methods~\cite{zoph2018learning,real2019regularized,liu2018darts,liu2018progressive,pham2018efficient,xu2019pc,liang2019darts+,chen2021neural} achieve affordable cost by only searching two types of cell structures, \ie, a normal cell and a reduction cell, which are shared across all the layers for constructing a neural network.
However, those methods can only search on a small proxy dataset due to high memory cost.
Most recent NAS algorithms~\cite{tan2019mnasnet,tan2019mixconv, tan2019efficientnet, cai2018proxylessnas,howard2019searching, wu2019fbnet,dai2019chamnet, wan2020fbnetv2,dai2020fbnetv3, zhou2019neural} employ well handcrafted inverted residual blocks (IRB) with varying kernel sizes and expansion ratios as search space candidates.
Though offering good efficiency,  such a constrained search space severely limits the exploration capability of NAS algorithms to search for more powerful network structures.

In this work, we propose a simple and efficient method for automating the search space design in two steps.
First, we conduct a one-time searching for the optimal subspace from a full search space on the target dataset.
Using the search space obtained, a typical NAS algorithm can be applied to search for the final network architecture. 
In this way, we can not only minimize the human interference in the search space design but also 
improve the network performance by searching in a better space. 
A comparison of the design choices of our proposed method and previous SOTAs are listed in Table~\ref{tab:space_size_comparison}. 
More discussions are deferred to supplementary materials.

\begin{table}[t]
 \footnotesize
\caption{\textbf{Search space design choices comparison.} The number of design choices in AutoSpace eclipse the previous algorithms' search space. AutoSpace automatically finds an optimized subspace of comparable size to \cite{cai2018proxylessnas} and \cite{tan2019mnasnet}.  ``Layer Variety'' indicates  if the searching algorithms allow different cell structures at different model layers; ``\#~\text{Models (log)}'' denotes the log10 value of the total number of architectures included in the search space for a 21-layer model.}

\label{tab:space_size_comparison}
\setlength\tabcolsep{1.5mm}
\centering
\begin{tabular}{lccccc}
\toprule
\multirow{2}{*}{\bf Algo.}  
&{\bf Layer}
&{\bf Search on}
&{\bf Space}
&{\bf \# Models} \\
&{\bf Variety }
&{\bf ImageNet}
&{\bf Design}
&{\bf ($\log$)}
\\ \midrule
DARTS~\cite{liu2018darts}  & \xmark  & \xmark & Manual & 2.38 \\
ENAS~\cite{pham2018efficient} & \xmark & \xmark & Manual & 3.70\\
PNAS~\cite{liu2018progressive} & \xmark & \xmark & Manual &5.74\\
Amoeba~\cite{real2017large} & \xmark & \xmark & Manual&5.74\\
NASNet~\cite{zoph2018learning}  & \xmark  & \xmark & Manual &7.85\\
MNasNet~\cite{tan2019mnasnet}& \cmark & \xmark & Manual & 52.72 
\\
SPOS~\cite{guo2019single}& \cmark & \cmark & Manual & 12.64 
\\
ProxylessNAS~\cite{cai2018proxylessnas} & \cmark & \cmark & Manual & 17.74 
\\
AutoSpace (ours) & \cmark & \cmark & Auto & 104.37 
\\ \bottomrule
\end{tabular}
\end{table}

\section{Method} 

\label{sec:method}

\begin{figure*}[t]
    \centering
    \includegraphics[width=0.95\textwidth]{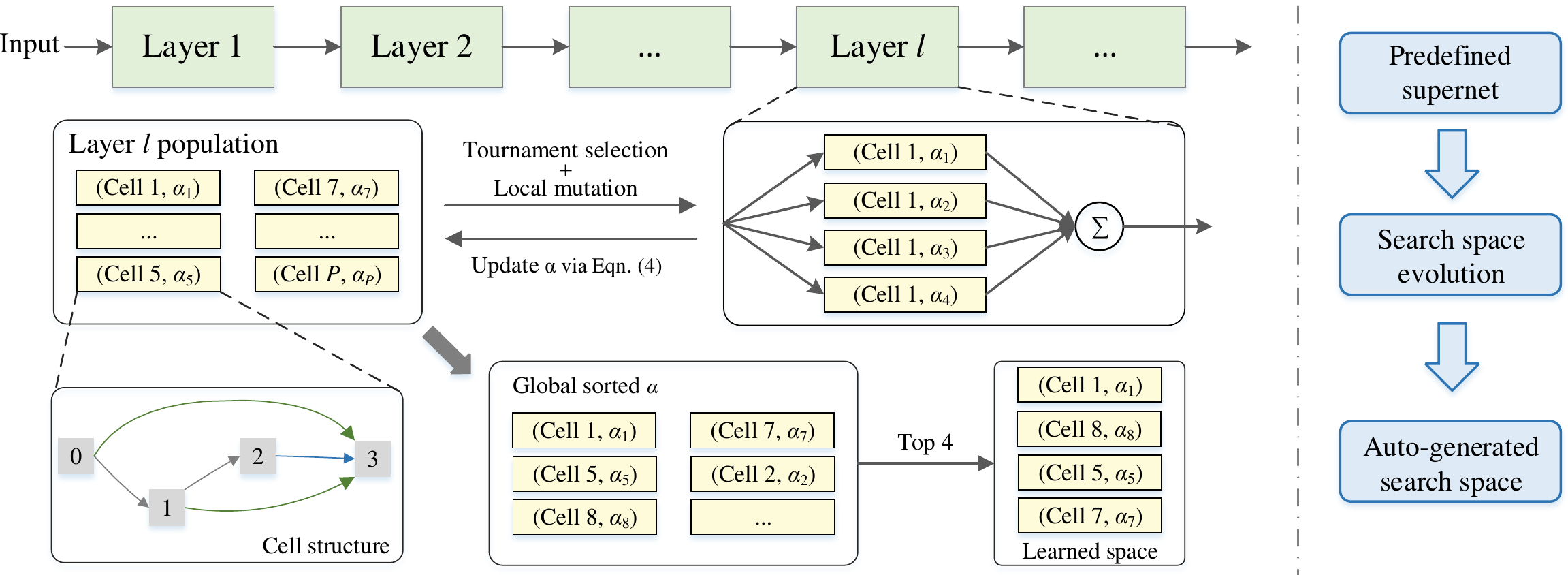}
    \caption{
    \textbf{Illustration on the search space generation process of AutoSpace}.
    Given a target network with a specified layer number and channel dimension, AutoSpace aims to learn an optimal layer-wise search space to be used in following NAS algorithms. 
    The entire process is performed by iterative sampling-and-update. 
    For each layer $l$, we first sample $K$ cells ($K=4$ in this example) from a randomly initialized population through tournament selection \cite{goldberg1991comparative, fang2010review}.
    $\alpha$ is the fitness score for each cell.
    Then a layer is  {represented by} stacking $K$ cells in parallel. The layer output is calculated as the weighted sum of the outputs of each single cell whthin the layer (ref. Eqn.~(\ref{eqn:motif_output})).
    In this way, the network can be trained on the target dataset with $\alpha$ updated efficiently via gradient back-propagation (ref. Eqn.~(\ref{eqn:grad_scaling})).
    In the end of each iteration, a cell's fitness score is updated via Eqn.~(\ref{eqn:fitness_score_update}).
    After the network converges, the top $K$ cells with the largest fitness scores in each layer are output as the final search space.}
    \label{fig:overall_flow}
\end{figure*}


\subsection{Problem Formulation}
\label{sec:formulation}

We consider the NAS problem of searching for a model architecture $N$ from a search space $S$ that consists of multiple basic building blocks (cells) $d$. 
Different from previous NAS work using a manually pre-defined search space, we aim to learn a proper search space automatically. 
We formulate the neural architecture search as a two-phase problem: 
first searching for an optimized search space and then searching network architectures using the optimized search space.

Formally, given an open search space $S$ with minimal human prior knowledge on the network architecture, we aim at finding a subset of the full space (called subspace) $S_{\mathrm{sub}}^\star \subset S$  which contains cells with optimized structures such that 
the constructed model architecture can achieve maximal accuracy $\mathrm{Acc}$
on the target dataset $\mathcal{D}$ at affordable computation cost (in MAdds). 
The objective can be formulated as a bi-level optimization problem:
\begin{equation}
\label{eqn:object_fnc}
\begin{aligned}
    {S^\star_{\mathrm{sub}}}  = & \mathop{\arg\max}_{S_{\mathrm{sub}} \subset S}  \text{Acc}({N}({d},S_{\mathrm{sub}}), \mathcal{D}), \\
    \text{s.t.} \quad  & d = \mathop{\arg\max}_{d\in S_{\mathrm{sub}}} \text{Acc}({N}({d},S_{\mathrm{sub}}), \mathcal{D}),
    \\ &  \text{MAdds}(d_i) < \text{MAdds}_{\max}, \forall d_i,
\end{aligned}
\end{equation}
where ${N}({d},S_{\mathrm{sub}})$ denotes the network searched  from the subspace $S_{\mathrm{sub}}$ and $d=\{d_1,d_2,\ldots\}$ denotes the set of selected cells for all the layers.
$\text{MAdds}_{\max}$ denotes the upper limit of the allowed MAdds of cell $d_i$. 
In particular, we consider searching for different subspaces at different layers of an $L$-layer model architecture, \ie, $S_{\mathrm{sub}} =  \prod_{i=1}^L S^i$ and $d_i \in S^i (i=1,\ldots,L)$.
In the following, we use $d_k^l$ to denote the $k^{th}$ candidate cell in the generated subspace of layer $l$, with $d_k^l \in S^l, |S^l|=K $.

Following~\cite{zoph2018learning,liu2018darts}, each cell in the subspace is represented by a directed acyclic graph (DAG) $G$.
Each node $x_i$ of $G$ is the output  feature representation from a certain edge and each directed edge ${G}_{ij}$ is associated with some operation
that transforms the input $x_i$ to $x_j$. 
In particular, $o$ denotes the set of basic operators defined below and  $o({G}_{ij})$ denotes the selected operation between node $i$ and $j$.

\myPara{Basic operators}
To minimize the human prior knowledge required on designing the network architecture, 
we do not specify any constraints on the node connection topology for the cells 
except for specifying two nodes as input and output of its DAG respectively. 
The edge connections are learned with our proposed learning framework
as detailed in Section~\ref{subsec:evolution}. 
For each edge, similar to most previous works \cite{cai2018proxylessnas,guo2019single,wu2019fbnet},
we consider five basic operations: $1\times 1$ convolution, $3 \times 3$ convolution,
depth-wise convolution, identity mapping, and the zero operation.
The input and output nodes are used as dimension adjustment nodes and following \cite{wu2019fbnet,cai2018proxylessnas}, the ratios between the input channels and the output channels $r \in \{1,3,6\}$ are learned through our proposed method which will be detailed in Section \ref{sec:ea_algo}.
The  aggregation function $o_f$ over the output features from different operators is selected from $\{\text{addition, dot product}\}$.

\subsection{AutoSpace Method}
\label{subsec:evolution}
Our AutoSpace learns the subspace in three steps as illustrated in Figure \ref{fig:overall_flow}. 
%
Following previous works~\cite{tan2019mnasnet,cai2018proxylessnas}, 
we construct an $L$-layer classification super-network (supernet)  with a $3\times 3$ convolutional layer as the head and one fully-connected layer for score prediction.
Each layer in between has $K$ parallel paths and each path is associated with  a candidate cell $d^l_k$ in the corresponding layer-wise search space $S^l$. 
The population of cells with different structures will be updated and evaluated via differentiable evolutionary algorithm (DEA).
After the evolution process, for each layer, the top-$K$ performing cells from the population are selected to form the search space $S^l$.

\subsubsection{Automatic search space generation}
\label{sec:ea_algo}
A typical evolutionary algorithm
consists of three steps \cite{back1993overview}. 
The first step is to generate a population of cells with different structures  (\ie,  different DAG connection topologies). 
We maintain a population of cells, denoted as $G^l$, for each layer of the constructed supernet.  
The second step is to apply a scoring function to evaluate the individual cell via tournament selection, where the winner cells with the largest $K$ fitness scores are allowed to generate off-springs via mutation. 
The third step is to generate new off-springs via mutation between the selected cells to enrich and improve the population.  
The last two steps are performed in an iterative manner. 
At the end of evolution, the top $K$ performing cells from the population will be used as the layer-wise subspace $S^l$.

However, the above EA method cannot be directly used to solve Eqn. (\ref{eqn:object_fnc}) and search for a subspace, due to the following challenges.
First, a large redundancy exits in the full space (population), incurring high computation overhead.
Secondly, a one-by-one evaluation for the subspace is extremely time consuming. 
Thirdly, an EA process usually starts  from scratch and is slow to converge.
To solve these challenges, we propose to leverage a reference DAG to assist in removing the redundant cells, speedup the convergence, and increase the parallelism for subspace evaluation.

\myPara{Reference DAG}  
Instead of evolving from scratch as convention, we propose to speedup the evolution process by inheriting the prior knowledge of a reference graph set at initialization.
We use ${G^{ref}}$ to denote the DAG of a verified well performing cell structure, and use it as the reference DAG.
Hamming distance is harnessed as a measure of similarity between DAGs~\cite{donnat2018tracking}. 
After each mutation, we calculate the hamming distance between the generated DAG and the reference DAG as follows:
\begin{equation}
\label{eqn:hamming_distance}
    r_H({G},{G^{ref}}) = \sum_{i,j} \frac{\vert{G}_{i,j} - {G^{ref}}_{i,j}\vert}{V(V-1)}.
\end{equation}
The mutation process will repeat until the distance is below the threshold $\tau$. 
Note that this reference graph regularization is only applied for the first half iterations of the entire training process to avoid adversely affecting learning of a better search space.

\subsubsection{Differentiable scoring function}

As our approach allows different layers to select different cell structures,
previous cell-based evaluation methods \cite{real2017large,real2019regularized} cannot be applied here due to the explosive increase in size of the search space,
and there is no one-to-one correspondence between network performance and quality of the cell structure.
To evaluate different cell structures, we propose to learn the fitness score $\alpha_k^l$ for each individual cell structure $d_k^l$  via gradient optimization to make the search process fully differentiable.

As illustrated in Fig.~\ref{fig:overall_flow}, we maintain a layer-wise population ${G}^l$ for layer $l$ of the supernet.
Each layer is initialized with $K$ randomly sampled cell structures 
$\{{G}^l_1,{G}^l_2,...,{G}^l_K\}$ from the corresponding population ${G}^l$.
The output of each sampled cell structure from the population will be weighted by its fitness score in the constructed supernet: 
\begin{equation}
\label{eqn:motif_output}
    f_{S^l} = \sum_{k=1}^K p_k d_k(x) = \sum_{k=1}^K \frac{\alpha_k^l}{\sum_j \alpha_j^l} d_k(x),
\end{equation}
%
 where $p_k$ is the weight for each cell structure $d_k$ in the supernet and $f_{S^l}$ is evaluated by computing its classification loss over the provided training dataset $\mathcal{D}$. 
 In this manner, multiple individuals at different layers can be evaluated concurrently. 
 To save the computation memory, we employ the binary gating function as proposed in \cite{cai2018proxylessnas} when learning the fitness score for each selected cell.
 %

The fitness scores for the selected $K$ cells in certain layer's population will be updated by gradient back-propagation when training the supernet to minimize the cross-entropy loss on the target dataset. 
To alleviate the imbalanced gradient updates on the fitness scores in each cell population due to the random  sampling in tournament selection,
we compensate the gradient update step of the fitness score via a scaling factor based on the training iteration number:
\begin{equation}
\label{eqn:grad_scaling}
    \frac{\partial L}{\partial \alpha_i^l} \approx \sum_{k=1}^K \frac{\partial L}{\partial g_k} p_k (\delta_{i,k} - p_i) \frac{n(d_i^l)}{n'(d_i^l)},
\end{equation}
where $g_k$ is the binary gates as introduced in \cite{cai2018proxylessnas} and $\delta_{i,k} = 1$ if $i=k$ and $ 0$ otherwise.  
$n(d_i^l)$ denotes the accumulative training iteration number in the supernet and $n'(d_i^l)$ denotes that for $d_i^l$ in the whole population. 
More discussions can be found in the supplementary materials.

After training the supernet for a few iterations, the updated fitness scores in the supernet, denoted as $\alpha^{\star}$, will be used to update those of corresponding cell structures in the population as below:
 \begin{equation}
\label{eqn:fitness_score_update}
    {\alpha_k^l}^{(t)} = \epsilon {\alpha_k^l}^{\star} + (1-\epsilon) {\alpha_k^l}^{(t-1)},
\end{equation}
where $\epsilon \in [0,1]$ is the momentum hyperparameter for updating the fitness score, and $\alpha^{(t-1)}$ denotes the old  fitness score recorded in the population before the updates.

The supernet will be re-generated every ${f}$ iterations.
When sampling the new search space, the fitness scores and the cell structures are sampled in pairs.
The fitness scores for each selected cell structure in each population will be updated iteratively via Eqn.~(\ref{eqn:grad_scaling})  and Eqn.~(\ref{eqn:fitness_score_update}) until the supernet converges. 
After the evolution, the top-K cell structures at each layer will be selected to form the searched space. 
In this way, each training of the supernet will evaluate $K^L$ network structures concurrently and thus the evolution process is sped up by $K^ L$ times compared to sequential evaluation. 

\subsection{Architectures search on the generated space}
Our search space searching process only needs to be run once for a target dataset and the generated search space can be used by any searching algorithms. To verify the effectiveness of AutoSpace, we allow different cell structures for different layers and directly search on ImageNet dataset to remove the transfer-ability issue.
To speed up the training process and save computation memory, we implement the weights sharing scheme as introduced in ENAS \cite{pham2018efficient} such that each time we re-build the supernet, the weights are inherited from previous runs.
We select three previous SOTA architecture searching algorithms based on gradient optimization (G), reinforcement learning (RL) and random sampling (RS) methods respectively from ProxylessNAS (G and RL) \cite{cai2018proxylessnas} and SPOS (RS) \cite{guo2019single}. 
For both searching algorithms, we add MAdds as a regualrization term to the loss function following previous NAS works \cite{cai2018proxylessnas,tan2019mnasnet}:
\begin{equation}
\label{eqn:flops_reg_loss}
    Loss = Loss_{CE} + \lambda \text{MAdds}(N),
\end{equation}
where $Loss_{CE}$ is the cross entropy loss and $\text{MAdds}(N)$ is the number of MAdds operation times in the selected network.
As will be shown in experiments, under all the three different algorithms, our auto-generated search space outperforms \baseliness~ones significantly.



\section{Experiments} 
\label{sec:experiments}



\subsection{Setup}

%

\myPara{Searching algorithm} The learned search space from our method can be applied to various NAS algorithms. 
To evaluate the searched space, we use the differentiable searching method ProxylessNAS~\cite{cai2018proxylessnas} to search model architectures 
because it allows layer variety and supports direct model searching on large datasets with high computation efficiency.
Following \cite{cai2018proxylessnas,tan2019mnasnet}, we use MAdds-based regularization to control the computation cost of the searched models. 
Besides, to verify the generalization ability of our learned search space in different NAS algorithms, 
we also evaluate its performance with reinforcement learning based~\cite{cai2018proxylessnas} and random sampling based \cite{guo2019single} NAS algorithms.
More details can be found in our supplementary material.

\myPara{Model evaluation} We train the searched model on the ImageNet dataset \cite{russakovsky2015imagenet} 
with an initial learning rate of 0.1 and cosine learning rate decay policy for 250 epochs. 
The batch size and weight decay are set to 512 and $1e{-}4$, respectively.
We use 10 epochs for learning rate warmup. 
%
Following \cite{cai2018proxylessnas}, we do not apply extra data augmentation,
like AutoAugment~\cite{cubuk1805autoaugment,cubuk2020randaugment} for a fair comparison.
When comparing with state-of-the-art efficient models, we follow the same set of training hyper-parameters as them and use RandAugment \cite{cubuk2020randaugment}
with default hyper-parameters during training.

\myPara{Manually designed search spaces for comparison}
We choose the inverted residual block (IRB)~\cite{sandler2018mobilenetv2} based search space as the baseline for comparison, which has been adopted by many recent SOTA NAS algorithms~\cite{cai2018proxylessnas,tan2019mnasnet,tan2019mixconv, wu2019fbnet}. 
It includes six variants of IRB, \ie, IRB with kernel sizes of 3, 5, 7 and expansion ratios of 3, 6 respectively. It also includes a zero operator for skipping certain layers. Throughout the experiments, we refer to this baseline search space as ``\baseliness''.



\subsection{Ablation Study}

\myPara{Effectiveness of AutoSpace}
We first validate our proposed strategy, \ie, generating a subspace from the open space first and then searching for models within the subspace, via comparison with the following two popular NAS strategies. 
The first is to use the evolutionary algorithm to directly search for \emph{a model in the same open space} as ours, which consists of all possible combinations of basic operators;
the other is to search for \emph{a model in the manually-designed search space}. 

The performance of the searched models using the above two strategies on ImageNet 
are summarized in Table~\ref{tab:one_step_searching_comparision}.
As can be seen, the model searched by our method outperforms the baseline models significantly in terms of both Top-1 accuracy and computational efficiency, which delivers two interesting findings.
First, the superiority over the direct searching strategy demonstrates that our proposed strategy can generate a high-quality search space that enables the subsequent NAS algorithm to search for better network architectures. 
Second, compared with the handcrafted search space, our search space is optimized jointly 
with the model architecture on the target dataset.
This allows our method to learn better task-specific models and benefit from the end-to-end training pipeline.

\begin{table}[h]
\footnotesize
\centering
\caption{\small \textbf{Performance comparison of different searching strategies.} `S.S' denotes search space. 
`Full' denotes full search space as ours. `Manual' denotes \baseliness~search space.
}
\label{tab:one_step_searching_comparision}
\begin{tabular}{c@{\hskip 0.1cm}c@{\hskip 0.1cm}c@{\hskip 0.1cm}c@{\hskip 0.1cm}c} 
\toprule
\multicolumn{1}{c}{\bf S.S.}
&\multicolumn{1}{c}{\bf Search Algo.}  
&\multicolumn{1}{c}{\bf Param. (M)}
&\multicolumn{1}{c}{\bf MAdds (M)}
&\multicolumn{1}{c}{\bf Acc. (\%)}
\\ \midrule
Full & EA & 4.2 & 480 & 73.7 \\
Manual & RL  & 7.2 & 470 & 74.9  \\
AutoSpace & RL & 4.6 & $\bold{415}$ & $\bold{75.8}$  \\ \bottomrule
\end{tabular}
\end{table}

\myPara{Comparison to handcrafted search space}
To comprehensively compare our learned search space and the handcrafted one, 
we first run AutoSpace to generate layer-wise search spaces on ImageNet. 
Then, we run the ProxylessNAS gradient-based (G) searching algorithms \cite{cai2018proxylessnas}
with different  MAdds regularizations, obtaining models of sizes spanning 100-200M (100M+), 
300-400M (300M+), 400-500M (400M+), and 500-700M (500M+), respectively. 
%

From the results in Table \ref{tab:ss_ablation}, for each model size constraint, 
the searched model from our auto-generated search space always outperforms the one
from the manually-designed space by a large margin with the same searching algorithm. 
We argue that the advantage of our method is that our auto-generated search space 
is optimized for each layer on the target dataset, which is more likely to contain
superior model structures than the manually designed search space 
that includes only identical operators across layers.
Above results evidently validate the advantage of AutoSpace over the \baseliness~ search spaces. 



\begin{table}[!h]
\footnotesize
\caption{\small \textbf{Comparison among the searched models based on our search space and the \baseliness~(`manual') search space used by ProxylessNAS.} `Acc.' denotes the top-1 classification accuracy on ImageNet. }
\label{tab:ss_ablation}
\centering
\begin{tabular}{ccccc}
\toprule
\multicolumn{1}{c}{\bf  Budget}  
&\multicolumn{1}{c}{\bf Search Space}
&\multicolumn{1}{c}{\bf Param. (M)}
&\multicolumn{1}{c}{\bf MAdds (M)}
&\multicolumn{1}{c}{\bf Acc. (\%)   } \\
  \midrule 
\multirow{2}{*}{\makecell[l]{100M+ \\ MAdds}} 
& Manual~\cite{cai2018proxylessnas}  & 3.3 & 190 & 69.2 \\
&Ours & 3.3 & 175 & 71.1\\
\midrule 
\multirow{2}{*}{\makecell[l]{300M+ \\ MAdds}} 
& Manual~\cite{cai2018proxylessnas}  & 3.7 & 320 & 74.1 \\
&Ours & 4.3 & 340 & 75.3\\
\midrule 
\multirow{2}{*}{\makecell[l]{400M+ \\ MAdds}} 
& Manual~\cite{cai2018proxylessnas}  & 5.3 & 465 & 74.8 \\
&Ours & 4.7 & 430 & 75.7\\
\midrule 
\multirow{3}{*}{\makecell[l]{500M+ \\ MAdds}} 
& Manual~\cite{cai2018proxylessnas}  & 6.9 & 590 & 76.6 \\
&Ours & 6.4 & 570 & 77.0\\
&Ours & 7.2 & 650 & 77.2 \\
\bottomrule 
\end{tabular}
\end{table}

\myPara{Applications to different searching algorithms}
To verify the generalizability of the search space learned by AutoSpace, we run different searching algorithms on it as well as on the baseline search space to  compare the resulting models. 
We choose two representative searching algorithms which are based on reinforcement learning (RL) and random sampling (RS) respectively. 
For the RL searching algorithm, we choose the widely-used single path sampling method proposed in \cite{cai2018proxylessnas}.
For the RS algorithm, we use \cite{guo2019single}. 
The results are shown in Table~\ref{tab:searching_algo_comparison}. 
Under different computation budgets, the model searched using search spaces from AutoSpace significantly outperforms the ones from the  baseline search space. 
For example, in the group with 300M+ Madds, our model outperforms competing models searched by RL and RS by 0.8\% and 0.4\% respectively. These results further verify the superiority of our auto-learned search space.

\begin{table}[t]
\footnotesize
\setlength{\tabcolsep}{5pt}
\caption{\small \textbf{Performance of our auto-generated search space with different searching algorithms.} `Manual' denotes the \baseliness~ search space.  `RL' denotes  reinforcement learning  based algorithms as used in ProxylessNAS~\cite{cai2018proxylessnas}. `RS' denotes the random searching algorithm as used in SPOS~\cite{guo2019single}. `Cls' denotes the Top-1 classification accuracy on ImageNet.}
\label{tab:searching_algo_comparison}
\centering
\begin{tabular}{lcccc}
\toprule
\multicolumn{1}{c}{\bf Group}  
&\multicolumn{1}{c}{\bf Search  Algo.}
&\multicolumn{1}{c}{\bf Param. (M)}
&\multicolumn{1}{c}{\bf MAdds (M)}
&\multicolumn{1}{c}{\bf Cls (\%)} \\
\midrule 
\multirow{4}{*}{\makecell[l]{300M+}} 
& Manual-RL & 4.1 & 330 &  74.3 \\
& Ours-RL & 4.0 & 360 & 75.1 \\ \cmidrule{2-5}
& Manual-RS & 4.0 & 350 & 73.5 \\
& Ours-RS & 4.2 & 360 & 73.9 \\
\midrule 
\multirow{4}{*}{\makecell[l]{400M+}} 
& Manual-RL & 7.2 & 470 &  74.9 \\
& Ours-RL & 4.6 & 415 &  75.8\\ \cmidrule{2-5}
& Manual-RS & 4.3 & 420 &  74.4 \\
& Ours-RS & 4.6 & 430 &  74.9 \\
\bottomrule 
\end{tabular}
\end{table}


\subsection{Algorithm Analysis}

\myPara{Analysis on search space size} 
We use 6 nodes for each DAG including the two IO nodes. 
The IO nodes are only used to adjust the dimensions. 
Thus, there are totally 6 edge connections within each DAG for evolution. 
With the above definition, for an $L$ layer network, the total number of architectures included in each DAG is $5^6 \times 3 \times 2$ = 93750. 
As we allow a variety for different layers, the total number of networks that could be formed via those DAG is $93750^L$. 
A typical value of $L$ is 21~\cite{cai2018proxylessnas,tan2019mnasnet}, and using it as an example, the size of the full search space of AutoSpace is~$2^{325}\times$ larger than the DARTS search space. A more detailed comparison is shown in Table 1 in the main paper.

\begin{figure}[t]
    \centering
    \includegraphics[width=1.02\linewidth]{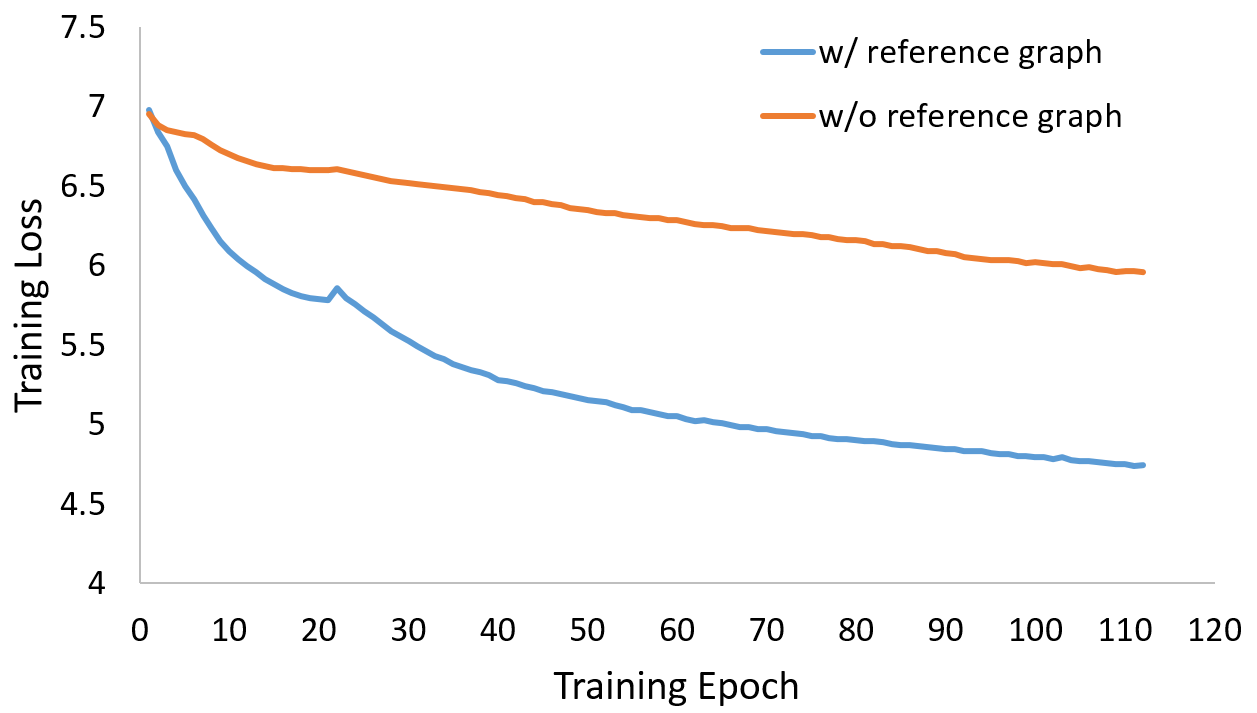}
    \caption{The training loss curve of the evolution process with and without reference DAG (blue vs.\ red line). The convergence is   faster with   the reference graph.
    }
    \label{fig:ref_graph_training}
\end{figure}

\myPara{Reference graph speedups convergence }
We further study the effects of using reference graph (described in Section~\ref{subsec:evolution}) for search space optimization. 
We use the reference graph regularization for the first 60 epochs and disable it for the rest of the evolution iterations.
The implementation of reference graph could speed up the convergence of the evolution of the supernet significantly as shown in Fig.~\ref{fig:ref_graph_training}. 
The overall searching time is reduced by $5.2 \times$ as detailed in Tab. \ref{tab:search_cost}. 

\myPara{Search Cost Analysis}
For the search space evolution process, we use 8 GPUs and run for 5 days. 
As we evaluate 9 cells for each fitness score update in Eqn. (\ref{eqn:fitness_score_update}),
there are in total $9^{21}$ different numbers of models in the constructed supernet. 
Thus, this evaluation approach is roughly $9^{21} \times$ faster than the conventional EA process with the same search space size. 
More importantly, the searching on the large space only needs to be run once on a given dataset and computation budget. A detailed comparison of searching cost with other NAS algorithms is shown in Table~\ref{tab:search_cost}.

\begin{table}[h]
\footnotesize
\caption{\small \textbf{Searching cost analysis on ImageNet.} `AutoSpace-G' denotes the searching cost on our learned search space with gradient optimization method proposed in \cite{cai2018proxylessnas}. The search space design only needs to be run once on the target dataset.
}
\label{tab:search_cost}
\begin{tabular}{lcc} 
\toprule
\multicolumn{1}{l}{\multirow{2}{*}{\bf Methods}}
&\multicolumn{1}{c}{\multirow{2}{*}{\makecell[c]{Search Space Design Cost \\ (GPU hours)}} }
&\multicolumn{1}{c}{\multirow{2}{*}{\makecell[c]{Searching Cost \\ (GPU hours)}} } 
\\ 
\\
\midrule
NasNet-A~\cite{zoph2018learning} & Manual &	48,000  \\
MNasNet~\cite{tan2019mnasnet} & Manual &	40,000   \\
AmoebaNet-A~\cite{real2019regularized} & Manual &	75,600   \\
ProxylessNAS~\cite{cai2018proxylessnas} & Manual &	200   \\
\midrule
AutoSpace-G (w/o ref) & 5000 &	200  \\ 
AutoSpace-G (w/ ref) & 960 &	200  \\ 
\bottomrule
\end{tabular}
\end{table}

\begin{table}[h]
\footnotesize
\centering
\caption{\small \textbf{Fitness score robustness of the differentiable scoring function.} 
We list the initial fitness scores for the three predefined networks. 
`Epochs' denotes the starting number of epochs for the fitness score, reflecting the correct ranking of the networks.
}
\label{tab:fitness_score_robotness}
\begin{tabular}{c@{\hskip 0.1cm}c@{\hskip 0.1cm}c@{\hskip 0.1cm}c@{\hskip 0.1cm}c} 
\toprule
\multicolumn{1}{c}{\bf S/N}
&\multicolumn{1}{c}{\bf $\alpha_{\text{Net 1}}$ ($\cdot 10^{-3}$)}  
&\multicolumn{1}{c}{\bf $\alpha_{\text{Net 2}}$ ($\cdot 10^{-3}$)}
&\multicolumn{1}{c}{\bf $\alpha_{\text{Net 3}}$ ($\cdot 10^{-3}$)}
&\multicolumn{1}{c}{\bf Epochs}
\\ \midrule
1. & -0.296 &	2.670 &	-0.141 & 5 \\
2. & 0.023 &	-0.577 &	0.705 & 3  \\
3. & 0.426 &	-0.752 &	-1.180 & 4  \\
4. & -1.950 &	-0.061 &	-0.334 & 5  \\
5. & 0.304 &	1.220 &	0.774 & 6  \\ 
\bottomrule
\end{tabular}
\end{table}

\begin{figure*}[h]
    \centering
    \includegraphics[width=1.0\textwidth]{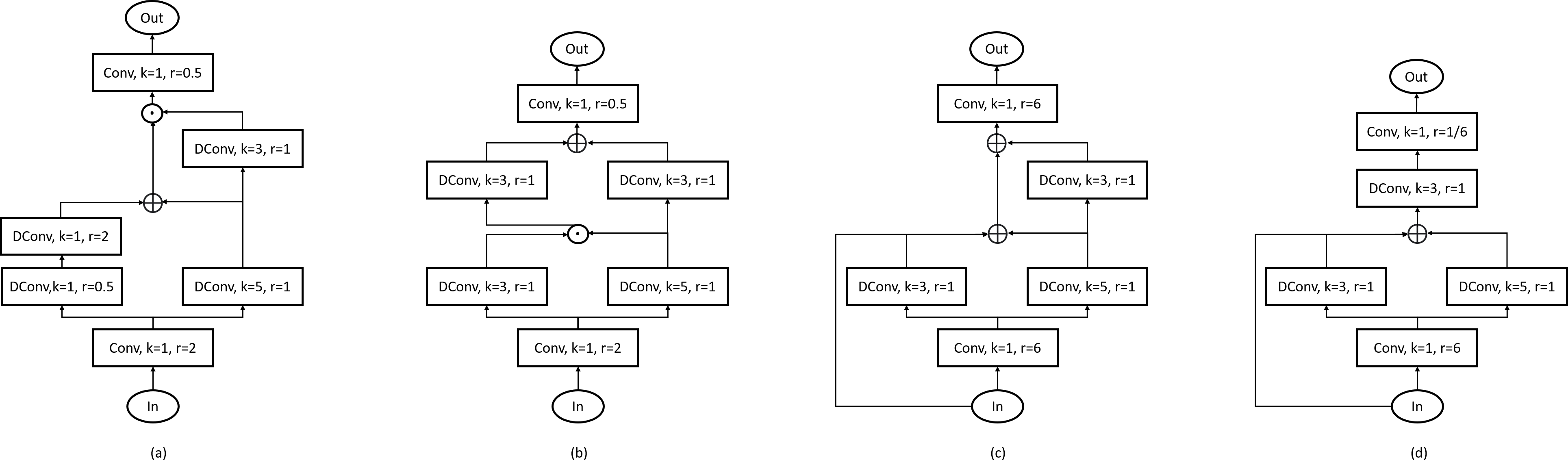}
    \caption{ \textbf{Illustration of the learned cell structures.} 
    We list four most frequently selected cell structures learned via AutoSpace.  `DConv' denotes depthwise convolution; `k' denotes the kernel size; `r' denotes the ratio between the output channel and the input channel of a convolution layer. `$\odot$' denotes dot product of two feature maps and `$\oplus$' denotes element-wise addition of two feature maps. Among the four cell structures, (d) is selected the most often.
    }
    \label{fig:learned_block}
\end{figure*} 

\myPara{Robustness to variations of fitness scores}
Each time we update the layerwise subspace with the differentiable evolutionary algorithm (DEA), 
the fitness score for each cell will be updated together with the cell structures 
based on the records in the population.
Thus, during training, it is expected that the fitness score can accurately 
reflect the relative advantages among selected cell structures.
To study the ranking accuracy of the differentiable fitness scoring function, 
%
%
%
%
following previous works~\cite{zhang2020deeper,yu2020evaluating}, we measure the stability of the generated rankings of three pre-defined networks with our proposed differentiable scoring functions and use this stability as a measurement of the robustness. 
Specifically, we manually design three networks with the basic building blocks proposed in ResNet~\cite{he2016deep}, denoted as Net1, Net2, and Net3. 
The three networks are using the same building block with different depth and the detailed configurations for these three networks can be found in our supplementary material. 
We first train the three networks on ImageNet to get the ground truth rankings of 
their classification accuracy: Net1 $<$ Net2 $<$ Net3.
Then, we use the proposed differentiable scoring function as introduced in Section~\ref{subsec:evolution}
to learn the fitness scores with different initial values.
The results are shown in Table~\ref{tab:fitness_score_robotness}.
Based on the results, with different initial values, the proposed differentiable scoring function is always able to learn the correct rankings of the three networks' fitness scores within 6 epochs.

\subsection{Comparison to the State-of-the-Arts}

\begin{table}[t]
\footnotesize
\caption{\textbf{Comparison of our proposed AutoSpace with previous SOTA NAS algorithms.} Our model  is obtained by applying ProxylessNAS  onto the auto-generated search space. For fair comparison, we split the methods into two groups, \ie, without SE~\cite{hu2018squeeze} or with SE (labeled with `\textdagger'). `$\star$' denotes the EfficientNet-b0 model improved with our searched cell structure as shown in Fig. \ref{fig:learned_block}(d).}
\label{sample-table}
\centering
\begin{tabular}{lcccc}
\toprule
\bf Method
&\bf Param. (M)
&\bf MAdds (M)
&\bf Top-1
\\ \midrule 
 MobileNetV2~\cite{sandler2018mobilenetv2} & 3.5 & 300 & 72.0 \\
 ShuffleNetV2-1.5~\cite{ma2018shufflenet}  & - & 299 & 72.6 \\
 DARTS~\cite{liu2018darts}  & 4.7 & 574 & 73.3 \\
 FBNet-C~\cite{wu2019fbnet}  & 5.5 & 375 & 74.9 \\
 ProxylessNAS-M~\cite{cai2018proxylessnas}  & 4.1 & 330 & 74.4 \\
 PNAS~\cite{liu2018progressive}  & 5.1 & 588 & 72.7 \\
 AmoebaNet-A~\cite{real2019regularized}  & 5.1 & 555 & 74.5 \\
 ChamNet-B~\cite{dai2019chamnet}  & 5.2 & 323 & 73.8 \\
 MNasNet-B1~\cite{tan2019mnasnet}  & 4.4 & 326 & 74.2 \\
 MobileNeXt~\cite{zhou2020rethinking} & 3.5 & 300 & 74.0 \\
\midrule
Ours &  4.3 & 340 & 75.3\\ \midrule[1pt]
MobileNeXt$^{\text{\textdagger}}$~\cite{zhou2020rethinking} & 6.1 & 590 & 76.1 \\
 MNasNet-A3$^{\text{\textdagger}}$~\cite{tan2019mnasnet}  & 5.2 & 403 & 76.7 \\
 EfficientNet-B0$^{\text{\textdagger}}$~\cite{tan2019efficientnet}  & 5.3 & 390 & 77.1 \\
 MixNet-M$^{\text{\textdagger}}$~\cite{tan2019mixconv}  & 5.0 & 360 & 77.0 \\
\midrule
Ours$^{\text{\textdagger}}$ & 5.7 & 380 & 77.5\\
Ours$^{\star}$ &  5.2 & 420 & 77.8\\
\bottomrule
\end{tabular}
\end{table}

\myPara{Classification on ImageNet}
We follow ProxylessNAS~\cite{cai2018proxylessnas} to use the single path searching algorithm for searching model architectures from the auto-learned search space. 
Then, we compare our model with other models obtained by manual design or popular NAS algorithms, the results of which are shown in Table~\ref{sample-table}.
For fair comparison, we divide compared models into two groups: models without SE modules~\cite{hu2018squeeze} and models with SE and extra data augmentation techniques. 
In each group, our model outperforms other models in terms of top-1 accuracy with less or comparable computation cost. 
As aforementioned, we believe this improvement comes from the high-quality cell structures learned with our proposed AutoSpace method. To further verify this, we summarize common characteristics appeared in the learned search space, as illustrated in Fig. \ref{fig:learned_block}. Among all the learned cell structures, we observe that (d) is the most frequently selected. To further investigate the performance of structure   (d), we replace all the DEA learned cells with (d) and add in SE modules~\cite{hu2018squeeze}  in the same manner as EfficientNet-B0~\cite{tan2019efficientnet}. Surprisingly, without using excessive data augmentation methods such as AutoAugmentation~\cite{cubuk1805autoaugment}, the modified model achieves 77.8\% Top-1  classification accuracy on ImageNet~\cite{krizhevsky2012imagenet}, outperforming the EfficientNet-B0 model by 0.7\%.

\section{Conclusion} \label{sec:conclusion}

In this work, we present the first affordable approach to generate search spaces for NAS   algorithms automatically, which targets at alleviating human effort in search space design. The search space generation is based on a substantially improved  evolutionary algorithm with three novel techniques to enhance the efficiency. Extensive experiments show that direct replacement of the manually-crafted search space with our auto-generated one could significantly improve performance of searched models in image classification. We believe the proposed method will motivate the research along the line of automatic search space design for NAS algorithms.


{\small
\bibliographystyle{ieee_fullname}
\bibliography{egbib}
}

\end{document}


\title{AutoSpace: Evolving Search Spaces for Neural Architecture Search\\Supplementary Material}

\author{First Author\\
Institution1\\
Institution1 address\\
{\tt\small firstauthor@i1.org}
\and
Second Author\\
Institution2\\
First line of institution2 address\\
{\tt\small secondauthor@i2.org}
}

\maketitle


\input{tex/appendix}

{\small
\bibliographystyle{ieee_fullname}
\bibliography{egbib}
}